\documentclass[manuscript,screen]{acmart}
\usepackage{amsmath}
\usepackage{graphicx}
\usepackage{booktabs} % For formal tables
\usepackage{tabularx}
\usepackage{enumitem}
\usepackage{xcolor}
\usepackage{soul}
\usepackage{placeins}
\usepackage{float}
\usepackage{balance}
\usepackage{adjustbox}

\usepackage{geometry}
\usepackage{adjustbox}
\usepackage{array}
\AtBeginDocument{%
  \providecommand\BibTeX{{%
    \normalfont B\kern-0.5em{\scshape i\kern-0.25em b}\kern-0.8em\TeX}}}

\acmBooktitle{}
\acmISBN{}

\begin{document}

%%
%% The "title" command has an optional parameter,
%% allowing the author to define a "short title" to be used in page headers.
\title{Controllable Topic-Focused Abstractive Summarization}

%%
%% The "author" command and its associated commands are used to define
%% the authors and their affiliations.
%% Of note is the shared affiliation of the first two authors, and the
%% "authornote" and "authornotemark" commands
%% used to denote shared contribution to the research.

\author{Seyed Ali Bahrainian}
\affiliation{%
  \institution{Computer Science Department, University of Tuebingen}
  \city{Tuebingen, Germany, email: bahrainian@brown.edu}
}

\author{Martin Jaggi}
\affiliation{%
  \institution{Computer Science Department, EPFL}
  \city{Lausanne, Switzerland, email: martin\_jaggi@epfl.ch}
}

\author{Carsten Eickhoff}
\affiliation{%
  \institution{Computer Science Department, University of Tuebingen}
  \city{Tuebingen, Germany, email: carsten\_eickhoff@brown.edu}
}

%%
%% By default, the full list of authors will be used in the page
%% headers. Often, this list is too long, and will overlap
%% other information printed in the page headers. This command allows
%% the author to define a more concise list
%% of authors' names for this purpose.
%\renewcommand{\shortauthors}{Trovato and Tobin, et al.}

%\renewcommand\Affilfont{\itshape\small}
%% THE ABOVE WAS COMMENTED OUT
%% The abstract is a short summary of the work to be presented in the
%% article.
\begin{abstract}
  Controlled abstractive summarization focuses on producing condensed versions of a source article to cover specific aspects by shifting the distribution of generated text towards a desired style, e.g., a set of topics. Subsequently, the resulting summaries may be tailored to user-defined requirements. This paper presents a new Transformer-based architecture capable of producing topic-focused summaries. The architecture modifies the cross-attention mechanism of the Transformer to bring topic-focus control to the generation process while not adding any further parameters to the model. We show that our model sets a new state of the art on the NEWTS dataset in terms of topic-focused abstractive summarization as well as a topic-prevalence score. Moreover, we show via extensive experiments that our proposed topical cross-attention mechanism can be plugged into various Transformer models, such as BART and T5, improving their performance on the CNN/Dailymail and XSum benchmark datasets for abstractive summarization. This is achieved via fine-tuning, without requiring training from scratch. Finally, we show through human evaluation that our model generates more faithful summaries outperforming the state-of-the-art Frost model. 
\end{abstract}

%%
%% The code below is generated by the tool at http://dl.acm.org/ccs.cfm.
%% Please copy and paste the code instead of the example below.
%%



\setcopyright{acmcopyright}
\acmJournal{TOIS}
\acmYear{2021} \acmVolume{1} \acmNumber{1} \acmArticle{1} \acmMonth{1} \acmPrice{15.00}\acmDOI{10.1145/3464299}

%%
%% Keywords. The author(s) should pick words that accurately describe
%% the work being presented. Separate the keywords with commas.
\keywords{sequence-to-sequence neural models, abstractive summarization, topical customization~\footnote{This article has some textual overlap with the PhD thesis of the first author~\cite{mythesis}.}}

%%
%% This command processes the author and affiliation and title
%% information and builds the first part of the formatted document.
\maketitle

Automatic document summarization produces a condensed version of a source document, covering its main aspects. Summarization systems are mainly classified into two categories: extractive summarization and abstractive summarization. 

Extractive summarization ~\citep{nallapati2017summarunner} \textit{selects} sentences of a source document based on a scoring scheme and combines those exact sentences to produce a summary. Conversely, abstractive summarization aims at producing shortened versions of a source document by \textit{generating} sentences that do not necessarily appear in the original document. The latter is receiving increased attention recently, due to how capable neural summarizers and pre-trained conditional language models \cite{lewis2019bart, raffel2019exploring} have become. The advent of sequence-to-sequence (seq2seq) architectures based on Long-Short-Term-Memory (LSTM) Networks (Hochreiter  and  Schmidhuber,  1997) with attention and pointer copy mechanism  ~\citep{nallapati2017summarunner, bahdanau2014neural, see2017get}, followed by Transformer architectures with multi-headed self-attention~\cite{vaswani2017attention}, significantly contributed to this trend.

\begin{figure}[t]
\includegraphics[width=0.45\textwidth]{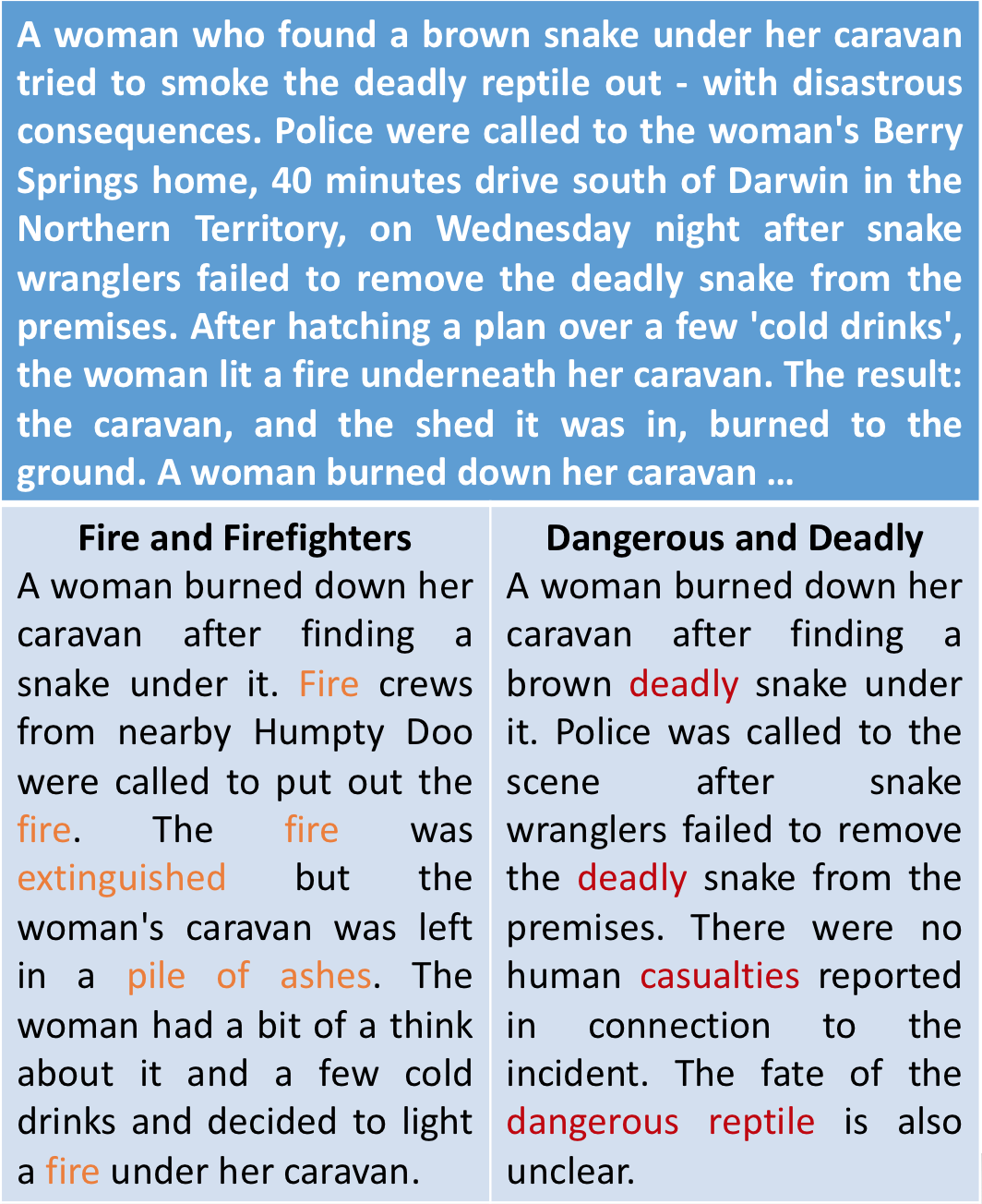}
\centering
 \caption{A news article (at the top) summarized with respect to two main topics underneath, Fire and Firefighters on the left, and Dangerous and Deadly on the right.}
  \label{fig:topicalExample}
\end{figure}

Transformer-based models~\citet{vaswani2017attention} generate high-quality summaries reaching  unparalleled  levels  of  fidelity, even outperforming human-written LEAD-3 baselines on summarization tasks~\cite{lewis2019bart, newts}. These models consist of an encoder that generates a rich representation of a source article, a decoder that generates summary tokens guided by the encoder representation, and learnable attention distributions that connect encoder and decoder to train the model end-to-end. However, the generation capabilities of these models may be controlled and directed towards producing text with a specific word distribution, e.g., focusing on topics of interest or even generating more faithful summaries~\cite{dou-etal-2021-gsum}.   

In this paper, we present a new abstractive summarization framework, featuring topic-focused text generation, adjusting the topics of a generated summary to conform to a target probability distribution over words, i.e.,~topics. A main component of this model is the topical cross-attention mechanism used to connect encoder and decoder guided by a topic model. We introduce the new architecture by adopting a \textit{Topical Attention} mechanism that can be easily integrated into any Transformer model. We name our model \textit{Conformer}, for CONtrolled topic-based transFORMER. It uses a topic model to compute distributions over words which, in turn, are learned by a feed-forward neural network to guide the cross-attention mechanisms of a summarization model to give more weight to specific user-defined topics when generating a summary. By using conditional generation (e.g., prompting) we show that the model further focuses its generated summaries on a target topic distribution. According to this definition, a topic is a probability distribution over the simplex (i.e., the vocabulary) such that the higher the associated probability of a word is, the higher is its presence in that topic. Therefore, the co-occurrence of words with a high probability forms a high-level semantic concept which may be the focus of a summary. Figure \ref{fig:topicalExample} illustrates an example document from the NEWTS dataset~\cite{newts} summarized twice, each time focusing on one of its main topics. We show that our model, Conformer, is not only highly effective in standard abstractive summarization, but also sets a new state-of-the-art performance on controlled topic-based summarization.

Latent Dirichlet Allocation (LDA) topic modeling by \citet{Blei:2003:LDA:944919.944937} follows the same co-occurrence notion, and is used to compute topics of corpora of textual documents. Conformer utilizes LDA for distinguishing topics. While Conformer is initialized using a pre-trained BART~\citep{lewis2019bart} model, we also demonstrate the ease of integrating our topic control mechanism into other Transformer models by also integrating it into T5 (Text-to-Text Transfer Transformer)~\citep{raffel2019exploring} model. Our experimental results show that Conformer sets a new state of the art on the XSUM benchmark dataset for abstractive summarization, as well as, on NEWTS for topic-based abstractive summarization..
%, while producing summaries using a richer, more diverse vocabulary. TODO

The main contributions of this paper are:
\begin{itemize}
    \item We present Conformer, a fundamentally new architecture for abstractive summarization based on a pre-trained conditional language model while requiring no additional complexity beyond the parameters of the original pre-trained model. 
    %We release our code-base to facilitate reproducibility of the results reported in this paper.
    \item We show that Conformer is highly effective at controlling the topical focus of a generated summary and demonstrate that the proposed architecture improves the faithfulness of the generated summaries.
\end{itemize}

The remainder of this paper is organized as follows: Section \ref{relatedwork} presents related work on abstractive summarization as well as the use of topic models~\cite{bahrainian-etal-2021-self-supervised} in summarization. In Section \ref{mainmodel}, we present our proposed topic-sensitive Transformer-based summarization model. In Section \ref{eval}, we evaluate our proposed approach comprehensively. Finally, Section \ref{conclusion} concludes the paper and presents an outlook on future directions.  

\section{Related Work}\label{relatedwork}
In this section, we discuss the body of related work on (1) attention-based abstarctive summarization models, (2) controlled text generation, and (3) the use of topic models in summarization:

\subsection{Attention in Abstractive Summarization}
One of the early deep learning architectures that was shown to be effective in abstractive summarization was the Attention-based Encoder-Decoder~\citep{nallapati2016abstractive} proposed by \citet{bahdanau2014neural} also known as vanilla seq2seq. Attention mechanisms have been shown to enhance the basic encoder-decoder model~\citep{bahdanau2014neural}. The main bottleneck of the basic encoder-decoder architecture was its fixed-sized representation (``thought vector''), which could not capture all the relevant information of the input sequence as the model or input scales up. However, the attention mechanism that relies on the notion that, at each generation step, only parts of the input are relevant has shown to be an effective solution to this problem. Following the vanilla seq2seq scheme, models such as the Pointer Generator Network (PGN)~\citep{vinyals2015pointer} with copy mechanism~\citep{see2017get} emerged in an effort to solve the challenge of out-of-vocabulary words. 

Later, Attention was shown to be effective in Transformer architectures~\cite{vaswani2017attention} as well. Several large pre-trained conditional language models for text generation were introduced. One such model is T5~\citet{raffel2019exploring} that can be fine-tuned for different seq2seq tasks. Another model in this category is BART~\citep{lewis2019bart}, a denoising autoencoder for pre-training seq2seq models. BART is trained by ``corrupting text with an arbitrary noising function and learning a model to reconstruct the original text''~\citep{lewis2019bart}. Our proposed model, Conformer, is initialized by a pre-trained BART model. In order to demonstrate the ease of integration of our proposed topical cross attention mechanism into arbitrary seq2seq Transformer models without adding new model parameters, we also plug this component into a T5 model and show the efficacy of this approach.

\subsection{Controlled Text Summarization}
Previous work has studied various control mechanisms to enforce a specific style~\cite{blinova-etal-2023-simsum}, form, or use of content in a generated summary.
\citep{DBLP:conf/emnlp/GehrmannDR18}~proposed to use a content selector to select phrases from a source document that should be part of a generated summary. Likewise,~\citep{DBLP:conf/emnlp/LiXLW18} introduce an information selection layer to explicitly model the information selection process in abstractive summarization. Other work has studied guiding the generation process based on topics~\cite{cats, Dathathri2020Plug}, aspects~\cite{https://doi.org/10.48550/arxiv.2210.02889}, entities~\cite{narayan-etal-2021-planning, maddela-etal-2022-entsum}, word relations such as triplets~\cite{dou-etal-2021-gsum} to shift the content of a generated summary toward a desired target form such as formal or informal text~\cite{DBLP:conf/naacl/BriakouLZT21}, biased versus neutral~\cite{DBLP:conf/aaai/PryzantMDKJY20}, simplified~\cite{cao-etal-2020-expertise}, or even sentimental stance~\cite{DBLP:conf/nips/ShenLBJ17}.

Here we study controlled text summarization using an information selection signal in the form of latent topics.

\subsection{Topic Models in Summarization}

%\carsten{Don't forget to cite CATS here.} TODO: is not published yet.

Previous work has utilized topic information in seq2seq problems such as neural response generation~\citep{cats,xing2017topic}. The work of \citet{xing2017topic} uses a topic model named Twitter LDA to respond to messages, a different objective than our work. \cite{sim-etal-2022-empirical} studied topic preservation in multi-document summarization. They conclude, that many summaries appear less informative due to a lack of topic coherence between them and their source articles. \cite{cui-hu-2021-topic-guided} incorporates a topic-based approach in a multi-document summary setting to jointly discover latent topics that can act as a semantic bridge across documents and provide global information to guide the summary generation. \cite{10.1007/978-3-030-86523-8_40} use a classifier to distill the training of a summarization model with respect to topical consistency between an input document and a generated summary. They observe that this brings improvements to standard abstractive summarization. 

%, their work is different from ours in that firstly, Twitter LDA assumes the existence of only a single topic per document. This assumption may be valid for tweet-length texts but will not hold in summarization of longer news articles. Secondly, their topic embeddings are derived from the source document, aggregated in a very different way from our approach, and are suitable only for their proposed architecture. In contrast, our approach relies on obtaining topics of the target summary and can be integrated into arbitrary Transformer-based models.

Another recent piece of related work integrates topic information into Transformer models for abstractive summarization \citep{wang-etal-2020-friendly}. The authors utilize the Poisson factor analysis~\citep{zhou2012beta} topic model. They add three new modules to a Transformer architecture at the cost of additional parameters. These three modules are semantic-informed attention, topic embedding with masked attention, and document-related modulation, which together integrate topic information into a Transformer architecture. Our work differs from theirs in three crucial ways: (1) our design does not introduce any additional model parameters to the original Transformer model, making it more efficient. (2) No training from scratch is required for the models in our approach, and standard fine-tuning for several epochs brings a significant improvement in summarization performance. (3) Our approach is more generic in that it can be integrated into various families of seq2seq models with attention and is not limited to Transformers.

%The use of LDA topic information in neural abstractive summarization has been considered by \citet{Wang:2018:RTC:3304222.3304389}. They use a reinforcement learning approach along with convolutional neural networks optimizing directly on ROUGE.

%\carsten{Presenting these methods in so much detail might make the reviewers curious about a side-by-side comparison. Are our results better than what they got and can we easily add an additional table in the results section that demonstrates that?}

In summary, topic information has been used in previous neural models as an input, and \citet{Wang:2018:RTC:3304222.3304389} argue that it results in the diversification of words appearing in summaries. However, the novelty of this paper lies in using this source of information to systematically influence the output summaries to conform to a specific topic distribution. This results in enhanced ROUGE scores for standard abstractive summarization while introducing a mechanism to control the topical distribution of generated summaries. Moreover, our approach is highly generic such that it can be easily integrated into existing seq2seq models at no additional computational cost (i.e., no new parameters are added to the Transformer model). At the same time, Conformer's novel topic-guided generation produces highly faithful summaries. 

\section{The Conformer Model}\label{mainmodel}

In this section, we introduce our new architecture featuring a topic-distribution adjusting mechanism capable of controlling the topics covered in the generated summaries. The model is equipped with a topical attention mechanism that can be easily plugged into any seq2seq Transformer-based model.

In the following, we first present the computation of topic-word probability distributions. Subsequently, we detail the integration of the topic information into the Transformer architecture.

\subsection{Computing Topic-word Distributions using LDA}

Although we use LDA as our topic model of choice, any other topic model that can factorize a corpus of text into two matrices, namely, the topic-word matrix and the document-topic matrix, may be used as a replacement. This requirement matches the properties of most existing topic models.

In order to compute the \textit{topical attention} weights, after training an LDA model on the training data, we map the target summary corresponding to each document to its LDA space. In other words, we compute the prevalence of all topics for each target summary, which results in a per-target summary probability distribution over all LDA topics. Furthermore, for each topic, LDA computes a probability distribution over words of the entire vocabulary~$\mathcal{V}$, i.e., the simplex. Therefore, for a given input document $d$ we can calculate a \textit{topical word vector} $\tau^d$ of dimension $\left |\mathcal{V} \right |$ considering all the words in that document, such that:
\begin{equation} \label{eq:topical_word_vectors}
\tau^d = \sum_{i} P(\text{topic}_{i}|d) \cdot \tilde{\mathbf w}_i  
\end{equation}

\noindent where $P(\text{topic}_{i}|d)$ is the probability of each LDA topic being present in the target summary, and $\tilde{\mathbf w}_i$ is the $\left |\mathcal{V} \right |$-dimensional vector consisting of the probabilities $\tilde{w}_{i,j} = P(\text{word}_j|\text{topic}_{i})$ of all words in vocabulary $\mathcal{V}$ under $\text{topic}_{i}$.

After having obtained the topic distribution of a target summary, we can now compute the probability scores of all tokens/words of the corresponding source document. In this way during training, all tokens/words from the source article will be scored according to the target topics that an abstractive summarization decoder needs to focus on when generating output summaries. That also means that the topic-words distribution is sorted and trimmed to match the source documents dimensions. In other words, each word from the source article is associated with a corresponding topic-words probability score computed using the LDA model. For stopwords or rare words that do not appear in the topic model vocabulary, a very small positive probability of `$10^{-9}$' is assigned.

The advantage of using topics of target summaries to guide the topical attention is that, if a word in the source document does not appear in any of the topics present in the target summary, this word will receive almost a `$0$' probability in the topic-words probability vector described in Equation \eqref{eq:topical_word_vectors}. 

By relying on topic information derived from the target summaries, we achieve two goals: (1) This focuses the summary generation attention precisely on the target topics as indicated by the target summaries, making our model a suitable solution for both standard and controlled abstractive summarization. This will direct the model to assign very low probabilities to tokens/words of the source document outside the scope of the corresponding target summary's topics. (2) According to previous research \citep{Wang:2018:RTC:3304222.3304389} the incorporation of topical information enables a language generation model to further diversify the vocabulary used in the generation process, thus generating a richer summary.

Due to the absence of target summary information at test time we train a feedforward neural network to map an input vector representing the frequency of words in an input article to its corresponding topic-words distribution computed by LDA defined in Equation \ref{eq:topical_word_vectors}. The feedforward network has one layer and with equal input and output size in the size of $\left |\mathcal{V} \right |$ optimizing the cross-entropy loss. The mapping of source article word frequencies to topic-words values is learned in the training phase when target summaries are available. Subsequently, the topic-words distribution computed by LDA is replaced by that of learned using the feedforward network both during training and testing to circumvent the absence of target summaries at test time. To elaborate, the feedforward network is trained based on LDA as described above, prior to fine-tuning the Summarizer model. From this point on $\tau^d$ refers to the topic-words distribution computed by the feedforward network. 
%Further details of the feedforward network is described in the appendix.

\subsection{Integrating Topic Information into Transformers}

This section discusses the integration of topical attention into the seq2seq Transformer architecture. Although we use a pre-trained BART model to initialize the Conformer, any summarization Transformer model may be used as well. In the experimental section, we demonstrate this on the example of a T5 model.  

\begin{figure}[t]
\includegraphics[width=8cm]{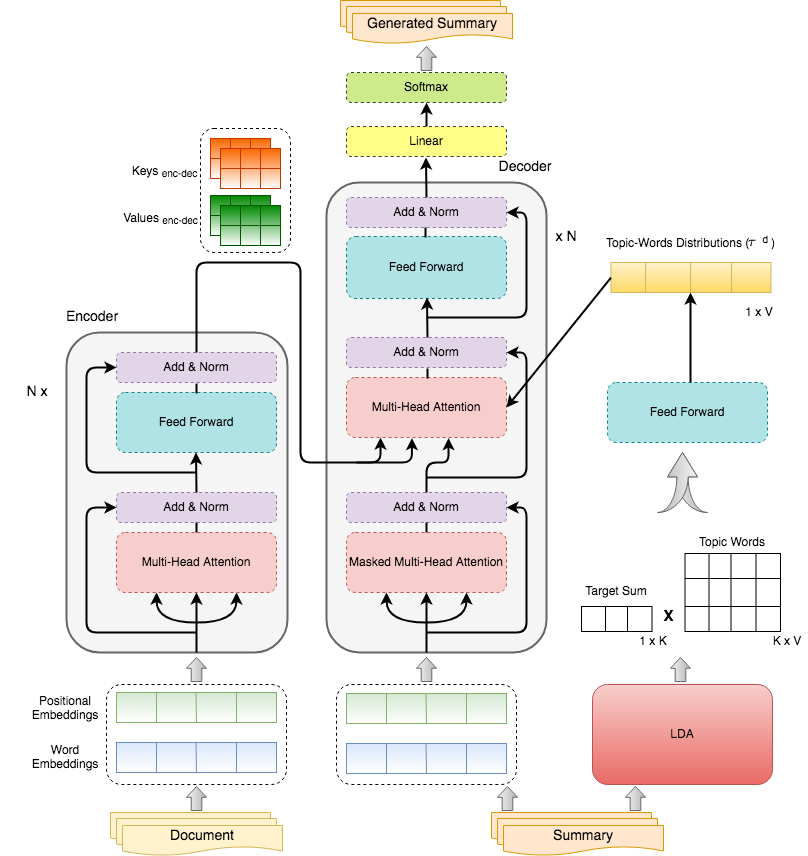}
\centering
 \caption{An illustration of Topical Attention in the Conformer Architecture. At training time, a target summary is projected to its LDA space. Equation \eqref{eq:topical_word_vectors} is calculated, yielding a $1 \times V$ vector, where $|V|$ is the vocabulary size. A feed-forward network learns and reconstructs the same vector. The vector is then reorganized in the size and order of input keys, and integrated into the Multi-head attention (in red color) using Equation~\eqref{transtopicalEq}.}
  \label{fig:transformer}
\end{figure}

Once we have calculated the topic probabilities of an input source document according to the topics of its corresponding target summary, we end up with a vector of topic probability values from the feedforward network in the size of $\left |\mathcal{V} \right |$. Subsequently, we trim and reorder this vector to match the size and the order of the input document. Figure \ref{fig:transformer} shows the integration of topical attention into the Transformer architecture which is shared by both BART and T5 models.

To integrate this vector into a Transformer architecture, we insert the topic-words probability scores derived from the output of the feedforward network into the cross-attention modules that connect the encoder with the decoder. That is, for each word of the source article, we compute a topic probability score, indicating the word's likelihood of appearing in the summary. Going back to the original Transformer paper~\citep{vaswani2017attention} each encoder layer has keys, queries, and values within the same layer. However, in the cross-attention module, the keys and values are computed by the encoder, while the decoder computes the queries.

This is because cross-attention essentially bridges the encoder blocks and the decoder blocks. Figure \ref{fig:transformer} shows how the key and value matrices flow from the encoder side to the decoder. 

While, for reasons of computational efficiency, the computations in a Transformer are performed at the matrix level, in the following, we present the steps of topical attention of the Conformer at the document level for illustrative clarity:

We remember from the original Transformer paper~\citep{vaswani2017attention} that cross-attention between encoder and decoder is computed as:
\begin{equation}\label{selfatten}
  \textit{Attention}(\textbf{Q},\textbf{K},\textbf{V}) = \text{softmax}\Big(\frac{\textbf{Q}\cdot \textbf{K}^\top}{\sqrt{d_{\textbf{K}}}}\Big)\textbf{V}
\end{equation}

The same equation holds in the case of cross-attention with the keys $K$ and values $V$ being outputs of the encoder while the queries $Q$ are computed on the decoder side. We observe from the equation that a dot product is computed between the keys and the queries. This dot product for a single source document and summary pair results in a single value indicating the importance of each key at its position given the query. Then this value is normalized, dividing it by $\sqrt{d_{K}}$ and then passing it to a softmax function. At this step, we compute an average between the result of the latter softmax function and the softmax of topic-words probabilities derived from Equation \eqref{eq:topical_word_vectors}. We, therefore, define topical attention as:
\begin{equation}\label{transtopicalEq}
\textit{Topical-Attention}(\textbf{Q},\textbf{K},\textbf{V}) =
 \tfrac12\bigg( \text{softmax}\Big(\frac{\textbf{Q} \cdot \textbf{K}^\top}{\sqrt{d_{K}}}\Big) + \text{softmax}({\tau}^d)\bigg)\textbf{V}
\end{equation}

Topical attention effectively reduces the importance of a word from the source document that is not covered in the topics discussed in the respective target summary by adding a small value or even `$0$' to it and then dividing it by $2$. Thus, the resulting value will become too small to be the focus of the generation process. On the other hand, the probability of a word related to a topic discussed in the target summary can be potentially increased or remain unchanged. We conclude that as explained above Conformer obtains topic information from the feed-forward network component, and is trained in an end-to-end fashion. 

%The other model components remain unchanged as reported in the respective papers on BART or~T5.

\section{Evaluation}\label{eval}
\subsection{Datasets}

We evaluate the effectiveness of all methods on three publicly available benchmark datasets.

\textbf{\textit{NEWTS}}~\cite{newts} is a topical abstractive summarization dataset that provides two alternative human-written summaries of each source article, with different topical focus for a subset of the CNN/dailymail collection. This dataset contains 4,800 training examples and 1,200 test samples. We use this dataset for evaluating controlled topical summarization.

The \textbf{\textit{CNN/DailyMail}} dataset~\cite{hermann2015teaching, nallapati2016abstractive}, contains news articles from the \textit{CNN} and \textit{Daily Mail} websites. The experiments reported in this paper are based on the non-anonymized version of the dataset, containing 287,226 pairs of training articles and summaries, 13,368 validation pairs, and 11,490 test pairs.

The \textbf{\textit{XSum}}~\cite{narayan-etal-2018-dont} dataset is included to show the efficacy of our approach on shorter news articles. This dataset is a collection of online articles from the British Broadcasting Corporation (BBC) and their corresponding summaries. It contains 204,045 training samples, 11,332 validation samples, and 11,334 test samples.

\subsection{Experimental Results}

\noindent\textbf{Topic-focused Summarization: } The primary feature of Conformer is controlling the topic distribution of the output summaries. Therefore, we first evaluate Conformer on the NEWTS dataset~\cite{newts} to measure the topical focus of the generated summaries with respect to target topics that should appear in a summary. This dataset presents each topic with four prompt types, namely, topic words, topic phrases, topic IDs and topic sentences. In this paper, we only use the sentence-based type to prompt our Conformer model to generate topic-focused summaries. Additionally, our model in this experiment is initialized based on the BART base model. At the same time, during the fine-tuning phase, the topical attention distribution enforces the presence of the target topics while limiting the rest. That is the weight of topics that are not the target of a summary are reduced to null. Table~\ref{benchmark} presents the results of this experiment. We report the baseline results from the original paper while using the same naming scheme. In the table, `b' following a model name indicates a `base' model size while `L' indicates a `large' model size. Additionally, `T-W' indicates the prompt `topic-words,' `T-ph' indicates a `topic-phrase' prompt, `T-Sent' indicates a `topic-sentence' prompt, `no prompt' means no prompting was used while fine-tuning a model, and `CNN-DM' indicates that the model was fine-tuned on the same source articles of our dataset paired with their original corresponding CNN/Dailymail summaries. In order to compute the topic focus score, we used the evaluation script from the Github repository of the NEWTS dataset~\cite{newts}. Additionally, we include results from a very large languge model, namely, ChatGPT (version: Aug 2023) to compare it against our model. In order to do so we use the prompt: \textit{Summarize this article with respect to topic: `T-Sent'}. We observe that our model defines a new state of the art both in terms of ROUGE performance as well as the topic-focus score computed by an LDA model as explained in the original paper. For completeness, the topic-focus score computes the prevalence of each target topic appearing in a corresponding summary averaged over all generated summaries. Furthermore, we observe that a significantly smaller Conformer outperforms ChatGPT in the controlled topic-focused summarization task one the NEWTS dataset. Based on the results of this experiment we conclude that Conformer is highly capable of shifting and controlling the topics of the output summaries.

\begin{table}[h]
\centering
\small
\begin{tabular}{c|c|c|c|c}
                      & R1 & R2 & RL & Topic Focus \\ \hline

BART-b + T-W      & 31.14  &  10.46 &  19.94 & 0.1375\\ %\hline
BART-b + T-Ph     &  31.01  &  10.36 & 19.91  & 0.1454\\ %\hline
BART-b + T-Sent  & 30.38  &  09.70 & 19.48  &  0.1513\\ %\hline
BART-b T-ID         &  30.97 & 10.23  &  20.08 &0.1399 \\ %\hline
%BART-b + T-Pred         &  16.12   &  1.52  & 12.22 & \\ %\hline
BART-b no prompt    & 16.48  &  0.75 & 11.71  & 0.0080\\ %\hline
BART-b CNN-DM    & 26.23  &  7.24 & 17.12  & 0.1338 \\ %\hline
T5-b + T-W        &  31.78 & 10.83  &  20.54 & 0.1386\\ %\hline
T5-b + T-Ph      &  31.55 &  10.75 &  20.27 & 0.1426\\ %\hline
T5-b + T-Sent    &  31.40 &  10.37 & 20.35  & 0.1528\\ %\hline
T5-b + T-ID           &  31.44 & 10.64  & 20.06  & 0.1342\\ %\hline
%T5-b + T-Pred         & 31.47  & 10.17  & 20.01  & \\ %\hline
T5-b no prompt         & 30.98  & 10.19  &  20.23 & 0.1379 \\ %\hline
T5-b CNN-DM         & 27.87  & 8.55  & 18.41  & 0.1305\\ %\hline
T5-L + T-W        & 30.92  & 10.01  &  20.19 & 0.1598\\ %\hline
T5-L + T-Ph      &  31.40 &  10.50 &  20.27 &  0.1457\\ %\hline
T5-L + T-Sent    &  30.64 & 09.84  & 19.91  & 0.1462\\ %\hline
T5-L + T-ID           & 30.35  &  9.93 &  19.77 & 0.1335\\ %\hline
%T5-L + T-Pred         &   &   &   & \\ %\hline
T5-L no prompt         &  30.06 & 9.55  & 19.25  & 0.1366 \\ %\hline
T5-L CNN-DM         & 28.44  & 8.49  &  18.61 & 0.1286 \\ %\hline

ProphetNet + T-W        &  31.91 &  10.80  & 20.66  &  0.1362\\ %\hline
ProphetNet + T-Ph      &  31.56 & 10.35 & 20.17  & 0.1474 \\ %\hline
ProphetNet + T-Sent    & 31.40  & 10.03  &  20.02 & 0.1633 \\ %\hline
%ProphetNet + T-ID           &   &   &   & \\ %\hline
%Pegasus + T-Pred         &   &   &   & \\ %\hline
ProphetNet no prompt         &  30.22 & 9.67 & 19.27  & 0.1316 \\ %\hline
ProphetNet  CNN-DM         &  28.71 & 8.53  & 18.69  & 0.1295 \\  %\hline
PPLM         & 29.63  &  9.08 &  18.76 & 0.1482\\ %\hline
CATS         & 30.12  & 9.35  & 19.11  & 0.1519 \\ %\hline
ChatGPT         & 32.47  & 11.26  & 20.58  & 0.1573 \\ %\hline
%\textbf{Conformer-b}(Ours)\!\! & \textbf{32.24} & \textbf{11.13} & \textbf{21.11} & \textbf{0.1708}\\
\textbf{Conformer-b}(Ours)\!\! & \textbf{34.16} & \textbf{11.67} & \textbf{21.93} & \textbf{0.1759}\\
%\hline
\end{tabular}
\newline
\caption{Benchmark comparing our model to the baselines reported in the original NEWTS paper, using a 3-fold cross validation in terms of $F_{1}$ ROUGE 1, $F_{1}$\ ROUGE 2, and $F_{1}$ ROUGE L and the LDA topic-focus score.}\label{benchmark}
\end{table}

\noindent\textbf{Evaluating Topical Attention: }This section evaluates our proposed model both in terms of best overall summarization performance, as well as, shifting the topical focus of the generated summaries. Model parameters are presented in the appendix. Following standard practice, we evaluate our proposed model against the baseline methods in terms of $F_{1}$ ROUGE 1, $F_{1}$\ ROUGE 2, and $F_{1}$ ROUGE L scores using the official Perl-based implementation of ROUGE~\cite{lin2004rouge}.  

%It is noteworthy that all model sizes are kept the same for each pair of models, i.e., an original model and its topical version.
\begin{table}
\caption{Comparison between BART and T5 models with and without topical attention in terms of $F_{1}$ ROUGE metrics on the CNN/Dailymail dataset in percentage ($\%$). }
\small
\vspace{-6pt}
 %\begin{adjustwidth}{-1.5cm}{}
\begin{tabular}{|l|c|c|c|}
\hline
    Models                       & R 1 & R 2 & R L \\ \hline \hline
%CATS    (Ours)           &   38.01    &    16.35     &    34.87     \\ \hline
%\textbf{PGN+coverage (Topical)}  &   \textbf{41.22}    &     \textbf{17.98}  &  \textbf{37.39}      \\ \hline
%PGN+coverage \cite{see2017get}) &     39.53      &    17.28    &  36.38       \\ \hline 
%\hline

%\textbf{Vanilla Attn. (Topical)} &    \textbf{36.56}     &     \textbf{14.16}    &  \textbf{33.35}      \\ \hline
%Vanilla Attn. \cite{nallapati2016abstractive} &    35.46     &     13.30    &  32.65      \\ \hline
%\hline

\textbf{T5-Topical-base (ours)}    &     \textbf{42.91}		      &     \textbf{20.64}   &  \textbf{39.88}    \\ \hline
T5-base \cite{raffel2019exploring}    &     42.05		      &     20.34   &  39.40    \\ \hline
\hline
\textbf{Conformer-base (ours)}     &     \textbf{43.79}     &     \textbf{21.17}   &  \textbf{40.77}     \\ \hline

BART-base \cite{lewis2019bart}     &     42.85     &     20.79   &  39.88     \\ \hline
\hline

\textbf{Conformer-large (ours)} &     \textbf{45.96}     &     \textbf{22.27}  &  \textbf{42.51}     \\ \hline
BART-large  \cite{lewis2019bart}      &     44.16     &     21.28   &  40.90     \\ \hline

\end{tabular}
\label{res1}
% \end{adjustwidth}
\end{table}

For each dataset an LDA model is trained followed by a feedforward network reconstructing the topic-words vector.
LDA returns $N$ probability distributions over the vocabulary representing the latent topics discussed in the respective training set. 
Since the actual number of underlying topics ($N$) is an unknown parameter in the LDA model, it needs to be estimated first. For this purpose, similar to the method proposed in~\cite{griffiths2004finding, Bahrainian:2018:AHM:3176349.3176399}, we perform a model selection process. It involves keeping the LDA parameters (commonly referred to as $\alpha$ and $\eta$) fixed while assigning several values of $N$ and running the LDA model for each value. We pick the model that minimizes the negative
$\log P (V | N )$, where $V$ contains all the words in the vocabulary of all the documents in the training data. This process is repeated until we find the topic model with the optimal number of topics. The training of each LDA model on CNN/Dailymail takes nearly a day, so we could only repeat it for a limited number of $N$ values. concretely, we trained the LDA models with values $N$ ranging from $50$ to $300$ in increments of $50$, and the optimal value on the CNN/Dailymail and XSum datasets were found to be $250$ and $150$, respectively. Subsequently, the feed-forward network learns LDA output for each input document.  

% We note that our Transformer-based models fine-tune a respective pre-trained model. 

Tables \ref{res1} and \ref{res2} present the results of this experiment. For all models and metrics, we can note superior performance of the Conformer on both the CNN/Dailymail and the XSum dataset compared against original models. 

Additionally, we observe that the relative performance improvement in all models against their corresponding original models is relatively higher in the case of the ROUGE 1 metric. One may interpret this observation such that the semantic similarity information obtained from the topic model forces a higher usage of individual topic-related words in the generated summaries. On the other hand, when it comes to ROUGE 2 and ROUGE L, since there are particular grammar (or word sequence) rules and patterns that a language model needs to enforce, the improvements are slightly smaller than those in ROUGE 1. \cite{dou-etal-2021-gsum} also states that generation guided by certain signals, e.g., in this case topicality, does improve the summarization performance.

It is worth noting that the demonstrated improvements in summarization performance come at no additional cost in terms of parameters added to the original models. This property of our scheme stands in stark contrast to other existing approaches such as \cite{wang-etal-2020-friendly} that require the addition of a significant number of model parameters. As an example, for the different Transformer models they study, they introduce an additional $5.69\%$ to $9.58\%$ of new parameters, which in absolute numbers amounts to 10.25 million and 38.91 million parameters according to their paper. Therefore, our approach is more generically integrated with Transformer-based models, and requires identical run time to the unmodified variants. Furthermore, our controlled generation approach is not language-specific and can be trained for any language.

\begin{table}
\caption{Comparison between BART and T5 models with and without topical attention in terms of $F_{1}$ ROUGE metrics on the XSum dataset in percentage~($\%$).}
\small
\vspace{5pt}
{\small
\begin{tabular}{|l|c|c|c|}
\hline
Models                       & R 1 & R 2 & R L \\ \hline \hline
\textbf{T5-Topical-base}     & \textbf{42.70} & \textbf{19.56} & \textbf{36.01} \\ \hline
T5-base \cite{raffel2019exploring} & 41.79 & 18.90 & 35.04 \\ \hline
\hline
\textbf{Conformer-base}      & \textbf{43.30} & \textbf{20.03} & \textbf{36.28} \\ \hline
BART-base \cite{lewis2019bart} & 42.32 & 19.14 & 35.52 \\ \hline
\hline
\textbf{Conformer-large}     & \textbf{47.67} & \textbf{24.95} & \textbf{39.81} \\ \hline
BART-large \cite{lewis2019bart} & 45.14 & 22.27 & 37.25 \\ \hline
\end{tabular}
}
\label{res2}
\end{table}

\begin{table*}
\centering
\caption{Comparison between our top-performing model against top recent baselines in terms of $F_{1}$ ROUGE metrics on the CNN/Dailymail and XSum datasets. Results cited from original papers.}
\small
\vspace{6pt}
\begin{tabular}{|l|c|c|c|c|c|c|}
\hline
\multicolumn{1}{|c|}{Datasets} & \multicolumn{3}{c|}{CNN/Dailymail} & \multicolumn{3}{c|}{XSum} \\
\hline
\multicolumn{1}{|c|}{Models} & \multicolumn{1}{c|}{R 1} & \multicolumn{1}{c|}{R 2} & \multicolumn{1}{c|}{R L} & \multicolumn{1}{c|}{R 1} & \multicolumn{1}{c|}{R 2} & \multicolumn{1}{c|}{R L} \\
\hline \hline
% \hline
%     Models                       & R 1 & R 2 & R L & R 1 & R 2 & R L \\ \hline \hline
%LEAD-3 Baseline  &   40.34   &     17.70   &  36.57      \\ \hline  

%RL with Intra-Attention~\cite{paulus2017deep}  &     41.16       &     15.75    &  39.08       \\ \hline
%BottomUpSum~\cite{DBLP:conf/emnlp/GehrmannDR18} &     41.22     &    18.68    &  38.34        \\ \hline

%InformationSelection~\cite{DBLP:conf/emnlp/LiXLW18}  &     41.54       &     18.18    &  36.47       \\ \hline      

%UnifiedAbsExt~\cite{DBLP:conf/acl/SunHLLMT18} &     40.68      &    17.97    &  37.13      \\ \hline      

\hline

UniLM~\cite{dong2019unified}     &     43.33	     &     20.21   &  40.51     &     42.63	     &    19.10   &  33.13\\ \hline

T5-largest~\cite{raffel2019exploring}    &     43.52		      &     21.55   &  40.69    &   -  &  -   &  -   \\ \hline

BART-large~\cite{lewis2019bart}     &     44.16     &     21.28   &  40.90     &   45.14    &   22.27     &    37.25   \\ \hline

ProphetNet~\cite{yan2020prophetnet}     &     44.20	      &    21.17  &  41.30    &  -   &  -   &   -  \\ \hline

BART+TA~\cite{wang-etal-2020-friendly} &  44.47  &   21.39 &  41.32        &  45.76  &   22.68 &  38.03\\     \hline

GSum~\cite{dou-etal-2021-gsum} &  45.94  &   \textbf{22.32} &  42.48        &  45.40  &   21.89 &  36.67\\     \hline

BRIO~\cite{liu-etal-2022-brio} &     \textbf{47.78}     &     \textbf{23.55}  &  \textbf{44.57}    &    \textbf{49.07}    &  \textbf{25.59}      &  \textbf{40.40} \\ \hline

\textbf{Conformer-large (ours)} &     45.96     &     22.27  &  42.51    &    47.67    &  24.95      &  39.81 \\ \hline

\end{tabular}
\label{res4}
\end{table*}

\noindent\textbf{Comparison to State of the art: }Now that we have demonstrated the positive effect of topical attention and topic controlability, we compare our top performing model to approaches from the literature. 
\noindent\textbf{Baselines:} These top-performing baseline models are: UniLM~\cite{dong2019unified}, BART \cite{lewis2019bart}, T5 ~\cite{raffel2019exploring}, BART+TA~\cite{wang-etal-2020-friendly}, GSum~\cite{dou-etal-2021-gsum}, and BRIO~\cite{liu-etal-2022-brio}.

Table \ref{res4} presents the results on the CNN/DailyMail and the XSum dataset. This table directly cites the results reported in the original papers, where available. 
We observe that our model performs superior to the latest Transformer models such as ProphetNet, the standard BART model and \cite{wang-etal-2020-friendly} on both datasets. At the same time, the top-performing approach of \cite{wang-etal-2020-friendly} which is also based on BART-large adds 38.91 million new parameters to the architecture while our approach adds no further parameters to the BART-large model. Comparing our results with theirs, we conclude that Conformer is highly effective to the point that the massive addition of hidden states in \cite{wang-etal-2020-friendly} does not lead to improvements over our approach. Moreover, in addition to the better computational efficiency of our model, our approach is more generic in that it can be integrated with various architectures as demonstrated in the case of T5 models. Finally, we observe that Conformer achieves superior performance over GSum~\cite{dou-etal-2021-gsum} on XSUM dataset while matching GSum's performance on the CNN/Dailymail dataset on abstractive summarization. This is while, the new BRIO model with its contrastive loss training outperforms our model. Studying the initialization of Conformer based on the BRIO model remains as future work.
%The evaluation and statistical significance has been carried out using the official ROUGE package~\cite{lin2004rouge}.

\noindent\textbf{Output Summaries:} We also include sample output summaries of our proposed  Conformer model in Table 6 below. The shift in the topical distribution of the words generated by the model when focusing on the topic `Marriage' and when focusing on topic `Charges and Arrest' is evident in the two summaries on the right side of the table.

\begin{table*}[ht]
\caption{From left to right, the table presents a human-written summary with no topical focus from the CNN/Dailymail dataset, Conformer generated summary with no topical focus, Conformer generated summary focused on the topic of `Marriage', and Conformer generated summary focused on the topic of Charges and Arrest}\label{lessTopcis_ex}
\small
\vspace{6pt}
\begin{tabularx}{\linewidth}{XXXX}

\multicolumn{1}{c}{\textit{\textbf{Human-Written Summary}}} &  \multicolumn{1}{c}{\textit{\textbf{Conformer(General)}}} & \multicolumn{1}{c}{\textit{\textbf{Conformer(Marriage)}}}  & \multicolumn{1}{c}{\textit{\textbf{Conformer(Charges and Arrest)}}}   

\tabularnewline  \bottomrule

Liana Barrientos \textcolor{blue}{married} ten men in eleven years - even \textcolor{blue}{marrying} six of them in one year alone . All of her \textcolor{blue}{marriages} took place in New York state. Her first \textcolor{blue}{marriage} took place in 1999, followed by two in 2001, six in 2002, and her tenth \textcolor{blue}{marriage} in 2010. Barrientos allegedly described her 2010 nuptials as 'her first and only \textcolor{blue}{marriage}' She is reportedly \textcolor{blue}{divorced} from four of her ten husbands. The Department of \textcolor{red}{Homeland Security} was 'involved' in Barrientos' case, the Bronx district \textcolor{red}{attorney}'s office has said. 
           & 39-year-old Lianna Barrientos \textcolor{blue}{married} ten men in eleven years and \textcolor{blue}{married} six of them in one year alone.
 Authorities reportedly slapped her with offering a false instrument for filing \textcolor{red}{felony charges} in November 2014. According to the New York Times, she has \textcolor{blue}{divorced} at least four times. The Department of \textcolor{red}{Homeland Security} was 'involved' in Barrientos' case, the Bronx district attorney's office told the New York Times. According to her Facebook profile she is from the Dominican Republic and lives in Manhattan.
                                    & 39-year-old Lianna Barrientos \textcolor{blue}{married} ten men in eleven years. She \textcolor{blue}{married} the ten men from 1999 to 2010 and \textcolor{blue}{married} six of the men in 2002 alone. The Times reported that for her first three 2002 \textcolor{blue}{marriages}, Barrientos \textcolor{blue}{wed} Koridze on Valentine's Day, before she went on to \textcolor{blue}{marry} Goktepe approximately two weeks later, then soon tied the knot with Paharelau. She claimed her last \textcolor{blue}{marriage} to Salle Keita was her first and only \textcolor{blue}{marriage}. 
   & Authorities reportedly slapped Lianna Barrientos with offering a false instrument for filing \textcolor{red}{felony charges} in November 2014
She was \textcolor{red}{arrested} and \textcolor{red}{charged} in November 2014 for lying regarding her \textcolor{blue}{marriage} to Salle Keita. Barrientos has been \textcolor{red}{arrested} multiple times, including for \textcolor{red}{loitering}, \textcolor{red}{drug possession}, and \textcolor{red}{jumping a turnstile} and \textcolor{red}{trespassing}, according to the Daily News. Department of \textcolor{red}{Homeland Security} was 'involved' in Barrientos' case, the Bronx district \textcolor{red}{attorney}'s office revealed to the Times. 
 \tabularnewline \bottomrule
\end{tabularx}

\end{table*}

\noindent\textbf{Summary Faithfulness: }It has been previously demonstrated that when generation is guided by auxiliary signals such as relations between words, e.g., triplets~\cite{dou-etal-2021-gsum} or entities~\cite{narayan-etal-2021-planning}, models generate more faithful summaries with less hallucination. In this experiment, we investigate whether guiding the generation process by topics, which are in essence groups of related words, can generate more faithful summaries. Therefore, we randomly select $100$ summaries from the test set of the CNN/Dailymail dataset and study the effect. We conduct a human study using three human judges while providing them with outputs of our model, that of a standard BART-large, as well as, the state-of-the-art Frost~\cite{narayan-etal-2021-planning} in faithfulness. Similar to \cite{narayan-etal-2021-planning}, each summary is rated from $1$ to $5$, with $1$ corresponding to very unfaithful content, $3$ corresponding to 50-50, and $5$ indicating a very faithful summary. The results of this experiment show that our model, Conformer, receives an average score of 4.62 while Frost and BART-large score 4.39 and 4.18 respectively, in how faithful their summaries are to the original document. This demonstrates another advantage of our model, indicating that a more granular guided approach (i.e.~ topics versus entities) is superior in terms of faithfulness and being factually correct. In the future we will further investigate the proclivity for hallucinations when the model is prompted with various topic-prompts including those not appearing in the source article.

\section{Conclusions and Future Work}\label{conclusion}

In this article, we introduce Conformer, a new summarization model conforming its summaries to a target topic distribution. It is based on a novel topical attention mechanism that is sufficiently generic to be end-to-end integrated into most seq2seq models via cross attention. Our approach does not impose any additional model parameters. Therefore, it is computationally as efficient as the original models. For the same reason, fine-tuning Conformer is also as easy as the original BART model.
We show that the incorporation of topical attention offers significant improvement in terms of ROUGE scores. We show that our novel Conformer model achieves state-of-the-art performance in topic-based abstractive summarization on the NEWTS dataset. This solution is not limited to a specific language by nature and can be trained for any languages. Finally, we showed that our approach yields summaries outperforming state-of-the-art in faithfulness. % while producing summaries with improved qualitative properties. \carsten{We do not show that.}
In the future, we plan to study controlled text generation to include/exclude various informative signals other than topics. Finally, we will investigate incorporating various user preference signals in the generation process.%examine the effectiveness of topical attention for other seq2seq problems.

%\section{Limitations}
%We discuss the limitations of this work as follows:

%\begin{itemize}
%    \item Topic models are generally more noisy if trained with fewer and shorter documents, due to the limited context to learn from. These models may benefit from longer documents. However, in this work each model was trained only on the target dataset and did not utilize any other datasets. The effect of training topic models on larger datasets needs to be analyzed and studied in the future.
%(1) A topic model used in a topic-based abstractive summarization Transformer may benefit from a different objective function to improve performance further beyond the results reported in this paper. (2) The performance of these abstractive summarization models is very dataset-dependant and consequently language-dependant. Fine-tuning a seq2seq model for the summarization tasks, does require dedicated datasets.
%    (3) Although, the controlled generation mechanism of Conformer is by design not language-specific and can be trained for other languages, we did not demonstrate this feature. 
%\end{itemize}

%\section*{Acknowledgements}

% Entries for the entire Anthology, followed by custom entries
%\bibliographystyle{ACM-Reference-Format}
%\bibliography{sample-base}

\clearpage

\appendix

%\section{Example Appendix}
%\label{sec:appendix}

\section{Model Parameters}

For completeness we state that the model parameters we used for fine-tuning the Transformer-based variants are, $beam-size=4$, $length\_penalty=2.0$, $max\_length=180$, $min\_length=56$, $no\_repeat\_ngram\_size=3$, and $early\_stopping=True$. The feed-forward network recostructing the word-topic probabilities is trained for 15 epochs with a single layer in the size 300. The models were fine-tuned for 5 epochs on a Tesla V100 GPU.

\
%This is an appendix.

%% The next two lines define the bibliography style to be used, and
%% the bibliography file.
\bibliographystyle{ACM-Reference-Format}
\bibliography{sample-base}

%%
%% If your work has an appendix, this is the place to put it.

\end{document}